# Improving Connectionist Energy Minimization

**Gadi Pinkas**	PINKAS@CS.WUSTL.EDU
*Center for Optimization and Semantic Control, Washington University*
*AMDOCS Inc, 1611 Des Peres Rd., St Louis, MO 63131 USA*

**Rina Dechter**	DECHTER@ICS.UCI.EDU
*Department of Information and Computer Science*
*University of California, Irvine, CA 92717, USA*

## Abstract

Symmetric networks designed for energy minimization such as Boltzman machines and Hopfield nets are frequently investigated for use in optimization, constraint satisfaction and approximation of NP-hard problems. Nevertheless, finding a global solution (i.e., a global minimum for the energy function) is not guaranteed and even a local solution may take an exponential number of steps. We propose an improvement to the standard local activation function used for such networks. The improved algorithm guarantees that a *global* minimum is found in *linear* time for tree-like subnetworks. The algorithm, called *activate*, is uniform and *does not* assume that the network is tree-like. It can identify tree-like subnetworks even in cyclic topologies (arbitrary networks) and avoid local minima along these trees. For acyclic networks, the algorithm is guaranteed to converge to a global minimum from any initial state of the system (self-stabilization) and remains correct under various types of schedulers. On the negative side, we show that in the presence of cycles, no uniform algorithm exists that guarantees optimality even under a sequential asynchronous scheduler. An asynchronous scheduler can activate only one unit at a time while a synchronous scheduler can activate any number of units in a single time step. In addition, no uniform algorithm exists to optimize even *acyclic* networks when the scheduler is synchronous. Finally, we show how the algorithm can be improved using the cycle-cutset scheme. The general algorithm, called *activate-with-cutset* improves over *activate* and has some performance guarantees that are related to the size of the network's cycle-cutset.

## 1. Introduction

Symmetric networks such as Hopfield networks, Boltzmann machines, mean-field and Harmony networks are frequently investigated for use in optimization, constraint satisfaction and approximation of NP-hard problems (Hopfield, 1982, 1984; Hinton & Sejnowski, 1986; Peterson & Hartman, 1989; Smolensky, 1986; Brandt, Wang, Laub, & Mitra, 1988). These models are characterized by a symmetric matrix of weights and a quadratic energy function that should be minimized. Usually, each unit computes the gradient of the energy function and updates its own activation value so that the free energy decreases gradually. Convergence to a *local* minimum is guaranteed although in the worst case it is exponential in the number of units (Kasif, Banerjee, Delcher, & Sullivan, 1989; Papadimitriou, Shaffer, & Yannakakis, 1990).

In many cases the problem at hand is formulated as a minimization problem and the best solutions (sometimes the *only* solutions) are the global minima (Hopfield & Tank, 1985; Ballard, Gardner, & Srinivas, 1986; Pinkas, 1991). The desired algorithm is therefore one that manages to reduce the impact of shallow local minima, thus improving the chances of finding a global minimum. Some models such as Boltzmann machines and Harmony nets use simulated annealing to escape from local minima. These models asymptotically converge to a global minimum, meaning that if the annealing schedule is slow enough, a global minimum is found. Nevertheless, such a schedule is hard to find and therefore, in practice, these networks are not guaranteed to find a global minimum even in exponential time.

In this paper we look at the *topology* of symmetric neural networks. We present an algorithm that finds a global minimum for acyclic networks and otherwise optimizes tree-like subnetworks in linear time. We then extend it to general topologies by dividing the network into fictitious tree-like subnetworks using the cycle-cutset scheme.

The algorithm is based on the method of nonserial dynamic programming methods (Bertelé & Brioschi, 1972), which was also used for constraint optimization (Dechter, Dechter, & Pearl, 1990). There the task was divided between a precompilation into a tree structure via a tree-clustering algorithm and a run-time optimization over the tree.

Our adaptation is connectionist in style; i.e., the algorithm can be stated as a simple, uniform activation function (Rumelhart, Hinton, & McClelland, 1986; Feldman & Ballard, 1982) and it can be executed in parallel architectures using synchronous or asynchronous scheduling policies. It does not assume the desired topology (acyclic) and performs no worse than the standard local algorithms for all topologies. In fact, it may be integrated with many of the standard algorithms in such a way that the new algorithm out-performs the standard algorithms by avoiding a certain class of local minima (along tree-like subnetworks).

Our algorithm is also applicable to an emerging class of greedy algorithms called *local repair algorithms*. In local repair techniques, the problem at hand is usually formulated as a minimization of a function that measures the distance between the current state and the goal state (the solution). The algorithm picks a setting for the variables and then repeatedly changes those variables that cause the maximal decrease in the distance function. For example, a commonly used distance function for constraint satisfaction problems is the number of violated constraints. A local repair algorithm may be viewed as an energy minimization network where the distance function plays the role of the energy. Local repair algorithms are sequential though, and they use a greedy scheduling policy; the next node to be activated is the one leading to the largest change in the distance (i.e., energy). Recently, such local repair algorithms were successfully used on various large-scale hard problems such as 3-SAT, n-queen, scheduling and constraint satisfaction (Minton, Johnson, & Phillips, 1990; Selman, Levesque, & Mitchell, 1992). Since local repair algorithms may be viewed as sequential variations on the energy minimization paradigm, it is reasonable to assume that improvements in energy minimization will also be applicable to local-repair algorithms.

On the negative side, we show that in the presence of cycles, no uniform algorithm exists that guarantees optimality even under a sequential asynchronous scheduler. An asynchronous scheduler can activate only one unit at a time while a synchronous scheduler can activate any number of units in a single time step. In addition, no uniform algorithm exists to optimize even *acyclic* networks when the scheduler is synchronous. Those negative results involve conditions on the parallel model of execution and therefore are applicable only to the parallel versions of local repair.

The paper is organized as follows: Section 2 discusses connectionist energy minimization. Section 3 presents the new algorithm *activate* and gives an example where it out-performs the standard local algorithms. Section 4 discusses negative results, convergence under various schedulers and self-stabilization. Section 5 extends the approach to general topologies through algorithm *activate-with-cutset* and suggests future research. Section 6 summarizes and discusses applications.

## 2. Connectionist Energy Minimization

Given a quadratic energy function of the form:

$$E(X_1, ..., X_n) = -\sum_{i<j}^{n} w_{i,j} X_i X_j - \sum_{i}^{n} +\theta_i X_i.$$

Each of the variables $X_i$ may have a value of zero or one called the activation value, and the task is to find a zero/one assignment to the variables $X_1, ... X_n$ that minimizes the energy function. To avoid confusion with signs, we will consider the equivalent problem of maximizing the goodness function:

$$G(X_1, ..., X_n) = -E(X_1, ..., X_n) = \sum_{i<j} w_{i,j} X_i X_j + \sum_{i} \theta_i X_i \qquad (1)$$

In connectionist approaches, we look at the network that is generated by assigning a node ($i$) for every variable ($X_i$) in the function, and by creating a weighted arc (with weight $w_{i,j}$) between node $i$ and node $j$, for every term $w_{i,j} X_i X_j$. Similarly, a bias $\theta_i$ is given to unit $i$, if the term $\theta_i X_i$ is in the function. For example, Figure 1 shows the network that corresponds to the goodness function $E(X_1, ..., X_5) = 3X_2 X_3 - X_1 X_3 + 2X_3 X_4 - 2X_4 X_5 - 3X_3 - X_2 + 2X_1$. Each of the nodes is assigned a processing unit and the network collectively searches for an assignment that maximizes the goodness. The algorithm that is repeatedly executed in each unit/node is called the *activation function*. An algorithm is *uniform* if it is executed by all the units.

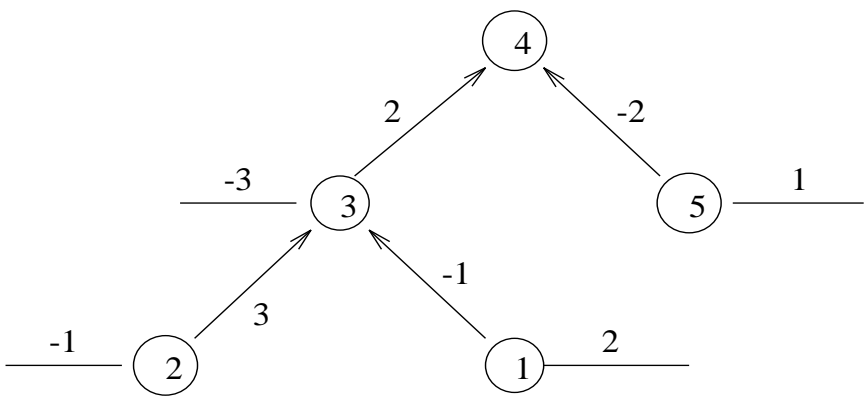

Figure 1: An example network

We give examples for two of the most popular activation functions for connectionist energy minimization: the discrete Hopfield network (Hopfield, 1982) and the Boltzmann machine (Hinton & Sejnowski, 1986). In the discrete Hopfield model, each unit computes its activation value using the formula:

$$X_i = \begin{cases} 1 & \text{if } \sum_j w_{i,j} X_j \geq -\theta_i \\ 0 & \text{otherwise} \end{cases}$$

In Boltzmann machines the determination of the activation value is stochastic and the probability to set the activation value of a unit to one is:

$$P(X_i = 1) = 1/(1 + e^{-(\sum_j w_{i,j} X_j + \theta_i)/T}),$$

where $T$ is the annealing temperature. Both approaches may be integrated with our topology-based algorithm; i.e., nodes that cannot be identified as parts of a tree-like topology use one of the standard local algorithms.

## 3. The Algorithm

We assume that the model of communication between neighboring nodes is a shared memory, multi-reader, single-writer model. We also assume (for now) that scheduling is done with a central scheduler (asynchronous) and that execution is fair. In a *shared memory, multi-reader, single-writer* each unit has a shared register called the activation register. A unit may read the content of the registers of all its neighbors but write only its own. *Central* scheduler means that the units are activated one at a time in an arbitrary order.[1] An execution is said to be *fair* if every unit is activated infinitely often. We do not require *self-stabilization* initially. Namely, algorithms may have an initialization step and can rely on initial values. Later we will relax some of the assumptions above and examine the conditions under which the algorithm is also self-stabilized.

The algorithm identifies parts of the network that have no cycles (tree-like subnetworks), and optimizes the free energy on these subnetworks. Once a tree is identified, it is optimized using a dynamic programming method that propagates values from leaves to a root and back.

Let us assume first that the network is acyclic; any such network may be directed into a rooted tree. The algorithm is based on the observation that given an activation value (0/1) for a node in a tree, the optimal assignments for all its adjacent nodes are independent of each other. In particular, the optimal assignment to the node's descendants are independent of the assignments for its ancestors. Therefore, each node $i$ in the tree may compute two values: $G_i^1$ is the maximal goodness contribution of the subtree rooted at $i$, including the connection to $i$'s parent whose activation is *one*. Similarly, $G_i^0$ is the maximal goodness of the subtree, including the connection to $i$'s parent whose activation value is *zero*. The acyclicity property will allow us to compute each node's $G_i^1$ and $G_i^0$ as a simple function of its children's values, implemented as a propagation algorithm initiated by the leaves.

Knowing the activation value of its parent and the values $G_j^0, G_j^1$ of all its children, a node can compute the maximal goodness of its subtree. When the information reaches the root, it can assign a value (0/1) that maximizes the goodness of the whole network. The assignment information propagates now toward the leaves. Knowing the activation value of its parent, a node can compute the preferred activation value for itself. At termination (at stable state), the tree is optimized. The algorithm has 3 basic steps:

1. **Directing a tree:** knowledge is propagated from leaves toward the center so that after a linear number of steps, every unit in the tree knows its parent and children.

2. **Propagation of goodness values:** the values ($G_i^1$ and $G_i^0$), are propagated from leaves to the root. At termination, every node knows the maximal goodness of its subtree and the appropriate activation value it should assign given that of its parent. In particular, the root can now decide its own activation value so as to maximize the whole tree.

3. **Propagation of activation values:** starting with the root, each node in turn determines its activation value. After O(depth of tree) steps, the units are in a stable state which globally maximizes the goodness.

---

1. Standard algorithms need to assume the same condition in order to guarantee convergence to a *local* minimum (Hopfield, 1982). This condition can be relaxed by restricting that only adjacent nodes are not activated at the same time (mutual exclusion).

Each unit's *activation register* consists of the following fields: $X_i$: the activation value; $G_i^0$ and $G_i^1$: the maximal goodness values; and $(P_i^1, .., P_i^j)$: a bit for each of the $j$ neighbors of $i$ that indicates $i$'s parent.

### 3.1 Directing a tree

The goal of this algorithm is to inform every node of its role in the network and its child-parent relationships. Nodes with a single neighbor identify themselves as leaves first and then identify their neighbor as a parent (point to it). A node identifies itself as a root when all neighbors point toward it. When a node's neighbors but one point toward it, the node selects the one as a parent. Finally, a node that has at least two neighbors *not* pointing toward it, identifies itself as being outside the tree.

The problem of directing a tree is related to the problem of selecting a leader in a distributed network, and of selecting a center in a tree (Korach, Rotem, & Santoro, 1984). Our problem differs (from general leader selection problems) in that the network is a tree. In addition, we require our algorithms to be self-stabilized. A related self-stabilizing algorithm appeared earlier (Collin, Dechter, & Katz, 1991). That algorithm is based on finding a center of the tree as the root node and therefore creates more balanced trees. The advantage of the algorithm presented here is that it is space efficient requiring only $O(log d)$ space, when $d$ is the maximum number of neighbors each node has. In contrast, the algorithm in Collin et al. requires $O(log n)$, $n$ being the network size.

In the algorithm we present, each unit uses one bit per neighbor to keep the pointing information: $P_i^j = 1$ indicates that node $i$ sees its $j$th neighbor as its parent. By looking at $P_j^i$, node $i$ knows whether $j$ is pointing to it.

Identifying tree-like subnetworks in a general network may be done by the algorithm in Figure 2.

In Figure 3a, we see an acyclic network after the tree directing phase. The numbers on the edges represent the values of the $P_i^j$ bits. In Figure 3b, a tree-like subnetwork is

---

**Tree Directing (for unit $i$):**

1. Initialization: If first time, then for all neighbors $j$: $P_i^j = 0$; /* Start with clear pointers (step is not needed in acyclic nets or with almost uniform versions) */

2. If there is only a single neighbor ($j$) and $P_j^i = 0$, then $P_i^j = 1$; /* A leaf selects its neighbor as parent if that neighbor doesn't point to it */

3. else, if one and only one neighbor ($k$) does not point to $i$ ($P_k^i = 0$), then $P_i^k = 1$, and for the rest of the neighbors: $P_i^j = 0$. /* $k$ is the parent */

4. else, for all neighbors $j$: $P_i^j = 0$. /* Node is either a root or outside the tree */

---

Figure 2: Tree directing algorithm

identified inside a cyclic network. Note that node 5 is not a root since not all its neighbors are pointing toward it.

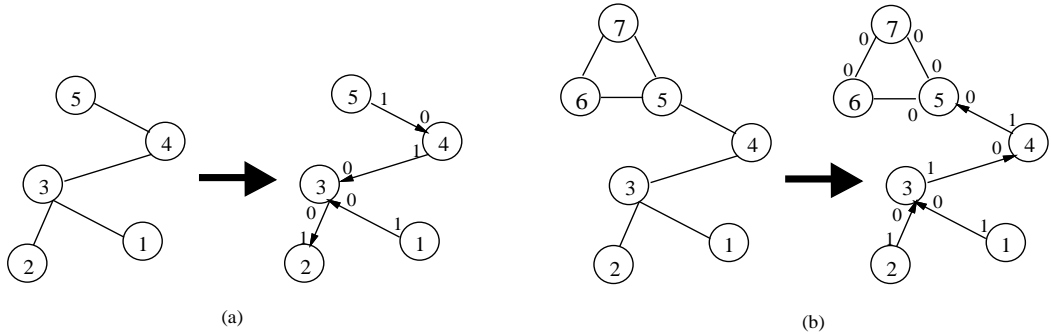

Figure 3: Directing a tree: a) A tree b) A cyclic network with a tree-like subnetwork.

## 3.2 Propagation of goodness values

In this phase every node $i$ computes its goodness values $G_i^1$ and $G_i^0$, by propagating these two values from the leaves to the root (see Figure 4).

Given a node $X_i$, its parent $X_k$ and its children, $children(i)$ in the tree, it can be shown, based on the energy function (1), that the goodness values obey the following recurrence:

$$G_i^{X_k} = max_{X_i \in \{0,1\}} \{ \sum_{j \in children(i)} G_j^{X_i} + w_{i,k} X_i X_k + \theta_i X_i \}$$

Consequently a nonleaf node $i$ computes its goodness values using the goodness values of its children as follows: If $X_k = 0$, then $i$ must decide between setting $X_i = 0$, obtaining a goodness of $\sum_j G_j^0$, or setting $X_i = 1$, obtaining a goodness of $\sum_j G_j^1 + \theta_i$. This yields:

$$G_i^0 = max \{ \sum_{j \in children(i)} G_j^0, \sum_{j \in children(i)} G_j^1 + \theta_i \}$$

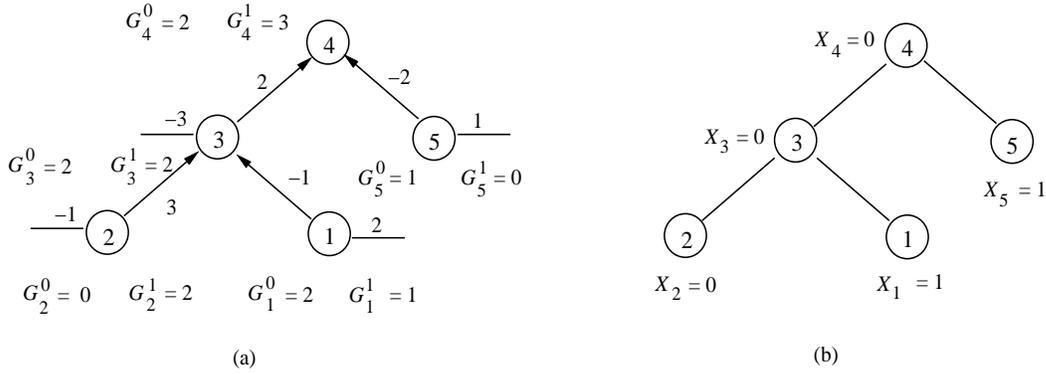

Figure 4: a) Propagating goodness values. b) Propagating activation values.

Similarly, when $X_k = 1$, the choice between $X_i = 0$ and $X_i = 1$, yields:

$$G_i^1 = max\{\sum_{j \in children(i)} G_j^0, \sum_{j \in children(i)} G_j^1 + w_{i,k} + \theta_i\}$$

The initial goodness values for leaf nodes can be obtained from the above (no children). Thus, $G_i^0 = max\{0, \theta_i\}$, $G_i^1 = max\{0, w_{ik} + \theta_i\}$.

For example, if unit 3 in Figure 4 is zero then the maximal goodness contributed by node 1 is $G_1^0 = max_{X_1 \in \{0,1\}}\{2X_1\} = 2$ and is obtained at $X_1 = 1$. Unit 2 (when $X_3 = 0$) contributes $G_2^0 = max_{X_2 \in \{0,1\}}\{-X_2\} = 0$ obtained at $X_2 = 0$, while $G_2^1 = max_{X_2 \in \{0,1\}}\{3X_2 - X_2\} = 2$ is obtained at $X_2 = 1$. As for nonleaf nodes, if $X_4 = 0$, then when $X_3 = 0$, the goodness contribution will be $\sum_k G_k^0 = 2 + 0 = 2$, while if $X_3 = 1$, the contribution will be $-3 + \sum_k G_k^1 = -3 + 1 + 2 = 0$. The maximal contribution $G_3^0 = 2$ is achieved at $X_3 = 0$.

Goodness values may be computed once for every node when its children's goodness values are ready; however, for self-stabilization (to be discussed later) and for simplicity, nodes may compute their goodness values repeatedly and without synchronization with their children.

### 3.3 Propagation of activation values

Once a node is assigned an activation value, all its children can activate themselves so as to maximize the goodness of the subtrees they control. When such value is chosen for a node, its children can evaluate *their* activation values, and the process continues until the whole tree is assigned.

There are two kinds of nodes that may start the process: a root which will choose an activation value to optimize the entire tree, and a non-tree node which uses a standard activation function.

When a root $X_i$ is identified, if the maximal goodness is $\sum_j G_j^0$, it chooses the value "0." If the maximal goodness is $\sum_j G_j^1 + \theta_i$, it chooses "1." In summary, the root chooses its value according to:

$$X_i = \begin{cases} 1 & \text{if } \sum_j G_j^1 + \theta_i \geq \sum_j G_j^0 \\ 0 & \text{otherwise} \end{cases}$$

In Figure 4 for example, $G_5^1 + G_3^1 + 0 = 2 < G_5^0 + G_3^0 = 3$ and therefore $X_4 = 0$.

An internal node whose parent is $k$ chooses an activation value that maximizes $\sum_j G_j^{x_i} + w_{i,k} X_i X_k + \theta_i X_i$. The choice therefore, is between $\sum_j G_j^0$ (when $X_i = 0$) and $\sum_j G_j^1 + w_{i,k} X_k + \theta_i$ (when $X_i = 1$), yielding:

$$X_i = \begin{cases} 1 & \text{if } \sum_j G_j^1 + w_{i,k} X_k + \theta_i \geq \sum_j G_j^0 \\ 0 & \text{otherwise} \end{cases}$$

As a special case, a leaf $i$ chooses $X_i = 1$ if $w_{i,k} X_k \geq -\theta_i$, which is exactly the discrete Hopfield activation function for a node with a single neighbor. For example, in Figure 4, $X_5 = 1$ since $w_{4,5} X_4 = 0 > -\theta_5 = -1$, and $X_3 = 0$ since $G_1^1 + G_2^1 + 2X_4 + \theta_3 = 1 + 2 + 0 - 3 = 0 < G_2^0 + G_1^0 = 2$. Figure 4b shows the activation values obtained by propagating them from the root to the leaves.

### 3.4 A complete activation function

Interleaving the three algorithms described earlier achieves the goal of identifying tree-like subnetworks and maximizes their goodness. In this subsection we present the complete algorithm, combining the three phases while simplifying the computation. The algorithm is integrated with the discrete Hopfield activation function demonstrating how similar the formulas are.

The steps of the algorithm can be interleaved freely; i.e., a scheduler might execute each step for all the nodes or all steps for any given node (or combinations). These steps are computed repeatedly with no synchronization with the node's neighbors.[2] Algorithm *activate* executed by unit $i$ (when $j$ denotes a non-parent neighbor of $i$ and $k$ denotes the parent of $i$) is given in Figure 5. Algorithm *activate* improves on an arbitrary local search connectionist algorithm in the following sense:

**THEOREM 3.1** *If $a_1$ is a local minimum generated by "activate" and $a_2$ is a local minimum generated by a local-search method (e.g., Hopfield), and if $a_1$ and $a_2$ have the same activation values on non-tree nodes, then $G(a_1) \leq G(a_2)$.*

*Proof:* Follows immediately from the fact that *activate* generates a global minimum on tree-subnetworks. □

Additional properties of the algorithm will be discussed in Section 4.

### 3.5 An example

The example illustrated in Figure 6 demonstrates a case where a local minimum of the standard algorithms is avoided. Standard algorithms may enter such local minimum and stay in a stable state that is clearly wrong.

The example is a variation on a Harmony network example (Smolensky, 1986) (page 259), and (McClelland, Rumelhart, & Hinton, 1986) (page 22). The task of the network is to identify words from low-level line segments. Certain patterns of line segments excite

---
2. As we will see later, the amount of parallelism will have to be limited somewhat, as from time to time two neighboring nodes should not execute the tree-directing step at the same time.

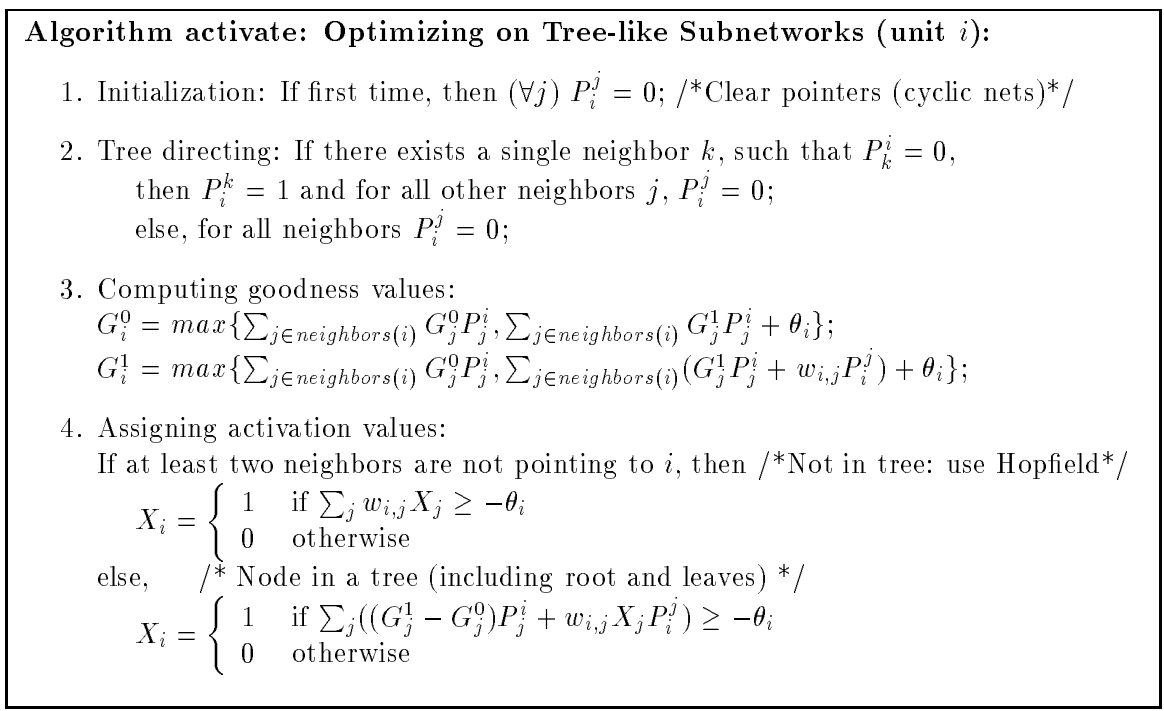

**Algorithm activate: Optimizing on Tree-like Subnetworks (unit $i$):**

1. Initialization: If first time, then $(\forall j)\ P_i^j = 0$; /*Clear pointers (cyclic nets)*/

2. Tree directing: If there exists a single neighbor $k$, such that $P_k^i = 0$,
   then $P_i^k = 1$ and for all other neighbors $j$, $P_i^j = 0$;
   else, for all neighbors $P_i^j = 0$;

3. Computing goodness values:
   $G_i^0 = max\{\sum_{j \in neighbors(i)} G_j^0 P_j^i, \sum_{j \in neighbors(i)} G_j^1 P_j^i + \theta_i\}$;
   $G_i^1 = max\{\sum_{j \in neighbors(i)} G_j^0 P_j^i, \sum_{j \in neighbors(i)} (G_j^1 P_j^i + w_{i,j} P_i^j) + \theta_i\}$;

4. Assigning activation values:
   If at least two neighbors are not pointing to $i$, then /*Not in tree: use Hopfield*/
   $X_i = \begin{cases} 1 & \text{if } \sum_j w_{i,j} X_j \geq -\theta_i \\ 0 & \text{otherwise} \end{cases}$
   else, /* Node in a tree (including root and leaves) */
   $X_i = \begin{cases} 1 & \text{if } \sum_j ((G_j^1 - G_j^0) P_j^i + w_{i,j} X_j P_i^j) \geq -\theta_i \\ 0 & \text{otherwise} \end{cases}$

Figure 5: Algorithm activate for unit $i$

units that represent characters, and certain patterns of characters excite units that represent words. The line strokes used to draw the characters are the input units: L1,..., L5. The units "N," "S," "A" and "T" represent characters. The units "able," "nose," "time" and "cart" represent words, and Hn, Hs, Ha, Ht, H1,... H4 are hidden units required by the Harmony model. For example, given the line segments of the character S, unit L4 is activated (input), and this causes units Hs and "S" to be activated. Since "NOSE" is the only word that contains the character "S," both H2 and the unit "nose" are also activated and the word "NOSE" is identified.

The network has feedback cycles (symmetric weights) so that ambiguity among characters or line-segments may be resolved as a result of identifying a word. For example, assume that the line segments required to recognize the word "NOSE" appear, but the character "N" in the input is blurred and therefore the setting of unit L2 is ambiguous. Given the rest of the line segments (e.g., those of the character "S"), the network identifies the word "NOSE" and activates units "nose" and H2. This will cause unit "N" and all of its line segments to be activated. Thus, the ambiguity of L2 is resolved.

The network is designed to have a global minimum when L2, Hn, "N," H2 and "nose" are all activated. However, standard connectionist algorithms may fall into a local minimum when all these units are zero, generating goodness of $5 - 4 = 1$. The correct setting (global minimum) is found by our tree-optimization algorithm (with goodness: 3-1+3-1+3-1+5-1-4+3-1+5=13). The thick arcs in the upper network of Figure 6 mark the arcs of a tree-like subnetwork. This tree-like subnetwork is drawn with pointers and weights in the lower part of the figure. Node "S" is not part of the tree and its activation value is set to one because the line-segments of "S" are activated. Once "S" is set, the units along the tree are optimized (by setting them all to one) and the local minimum is avoided.

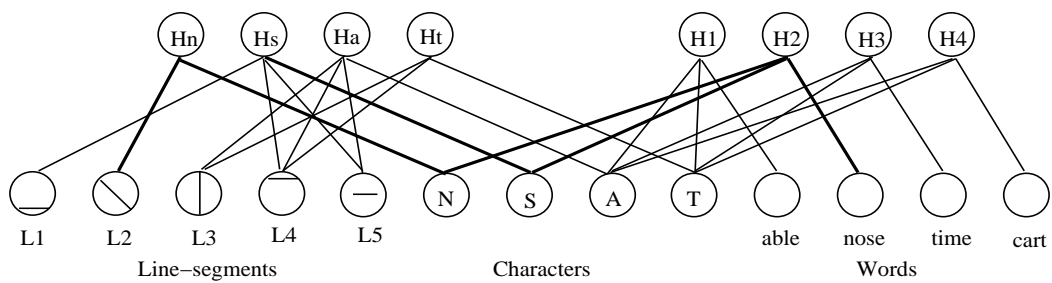

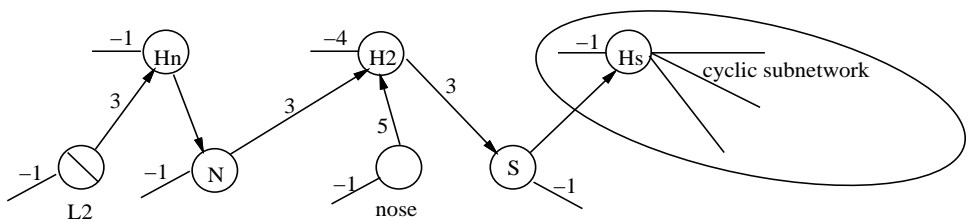

Figure 6: A Harmony network for recognizing words: local minima along the subtrees are avoided.

## 4. Feasibility, Convergence, and Self-Stabilization

So far we have shown how to enhance the performance of connectionist energy minimization networks without losing much of the simplicity of the standard approaches. The simple algorithm presented is limited in three ways, however. First, it assumes unrealistically that a central scheduler is used; i.e., a scheduler that activates the units one after the other asynchronously. The same results are obtained if the steps of the algorithm executes as one atomic operation or if neighbors are mutually excluded. We would like the network to work correctly under a *distributed (synchronous) scheduler*, where any subset of units may be activated for execution at the same time synchronously. Second, the algorithm guarantees convergence to global optima only for tree-like subnetworks. We would like to find an algorithm that converges to correct solutions even if cycles are introduced. Finally, we would like the algorithm to be *self-stabilizing*. It should converge to a legal, stable state given enough time, even after noisy fluctuations that cause the units to execute arbitrary program states and the registers to have arbitrary content. Formally, an algorithm is *self-stabilizing* if in any fair execution, starting from any input configuration and any program state (of the units), the system reaches a valid stable configuration.

In this section, we illustrate two negative results regarding the first two problems; i.e., that it is not feasible to build uniform algorithms for trees under a distributed scheduler, and that such an algorithm is not feasible for cyclic networks even under a central scheduler.

We then show how to weaken the conditions so that convergence is guaranteed (for tree-like subnetworks) in realistic environments and self-stabilization is obtained.

A scheduler can generate any specific schedule consistent with its definition. Thus, the central scheduler can be viewed as a specific case of the distributed scheduler. We say that a problem is *impossible* for a scheduler if for every possible algorithm there exists a fair execution generated by such a scheduler that does not find a solution to the problem. Since all the specific schedules generated by a central scheduler can also be generated by a distributed scheduler, what is impossible for a central scheduler is also impossible for a distributed scheduler.

### 4.1 Negative results for uniform algorithms

Following Dijkstra (1974), negative results were presented regarding the feasibility of distributed constraint satisfaction (Collin et al., 1991). Since constraint satisfaction problems can be formulated as energy minimization problems, these feasibility results apply also for computing the global minimum of energy functions. For completeness we now adapt those results for a connectionist computation of energy minimization.

THEOREM **4.1** *No deterministic[3] uniform algorithm exists that guarantees a global minimum under a distributed scheduler, even for simple chain-like trees, assuming that the algorithm needs to be insensitive to initial conditions.*

*Proof* (By counter example): Consider the network of Figure 7. There are two global minima possible : (11...1101...1) and (11...1011...1) (when the four centered digits are assigned to units, $i-1, i, i+1, i+2$). If the network is initialized such that all units have the same register values, and all units start with the same program state, then there exists a fair execution under a distributed scheduler such that in every step all units are activated. The units left of the center $(1, 2, 3, ...i)$ "see" the same input as those units right of the center $(2i, 2i-1, 2i-2, ..., i+1)$ respectively. Because of the uniformity and the determinism, the units in each pair $(i, i+1), (i-1, i+2), ..., (1, 2i)$ must transfer to the same program state and produce the same output on the activation register. Thus, after every step of that execution, units $i$ and $i+1$ will always have the same activation value and a global minimum (where the two units have different values) will never be obtained. □

This negative result should not discourage us in practice since it relies on an obscure infinite sequence of executions which is unlikely to occur under a random scheduler. Despite this negative result, one can show that algorithm *activate* will optimize the energy of tree-like subnetworks under a distributed scheduler if at least one of the following cases holds (see the next section for details):

1. If step 2 of algorithm *activate* in Section 3.4 is atomic; i.e., no other neighbor may execute step 2 at the same time.

2. if for every node $i$ and every neighbor $j$, node $i$ is executed without $j$ infinitely often (fair exclusion);

3. if one node is unique and acts as a root, that is, does not execute step 2 (an almost uniform algorithm);

---

3. The proof of this theorem assumes determinism and does not apply to stochastic activation functions.

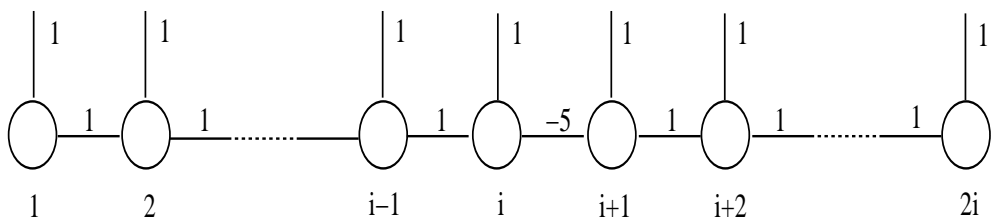

Figure 7: No uniform algorithm exists to optimize chains under distributed schedulers.

4. if the network is cyclic (one node will be acting as a root).[4]

Another negative result similar to (Collin et al., 1991) is given in the following theorem.

THEOREM 4.2 *If the network is cyclic, no deterministic uniform algorithm exists that guarantees a global minimum, even under a central scheduler, assuming that the algorithm needs to be insensitive to initial conditions.*

*Proof* (by counter example): This may be proved even for cyclic networks as simple as rings. In Figure 8 we see a ring-like network whose global minima are (010101) and (101010). Consider a fair execution under a central scheduler that activates the units 1,4,2,5,3,6 in order and repeats this order indefinitely. Starting with the same program state and same inputs, the two units in every pair of (1,4), (2,5), (3,6) "see" the same input, therefore they have the same output and transfer to the same program state. As a result, these units never output different values and a global minimum is not obtained. □

Note that any tree-like subnetwork of a cyclic network will be optimized even under a distributed scheduler (since nodes that are part of a cycle are identified as roots and the algorithm acts as an almost uniform algorithm).

## 4.2 Convergence and self-stabilization

In the previous subsection we proved that under a pure distributed scheduler there is no hope for a uniform network algorithm. In addition, we can easily show that the algorithm is not self-stabilizing when cycles are introduced. For example, consider the configuration of the pointers in the ring of Figure 9. It is in a stable state although clearly not a valid tree.[5]

In this subsection we weaken the requirements allowing our algorithm to converge to correct solutions and to be self-stabilizing under realistically weaker distributed schedulers. We will not use the notion of a *pure* distributed scheduler; instead, we will ask our distributed scheduler to have the *fair exclusion* property.

---

4. Global solutions are not guaranteed to be found but all tree-like subnetworks will be optimized.
5. Such configuration will never occur if all units start at the starting point; i.e., clearing the bits of $P_i$. It may only happen due to some noise or hardware fluctuations.

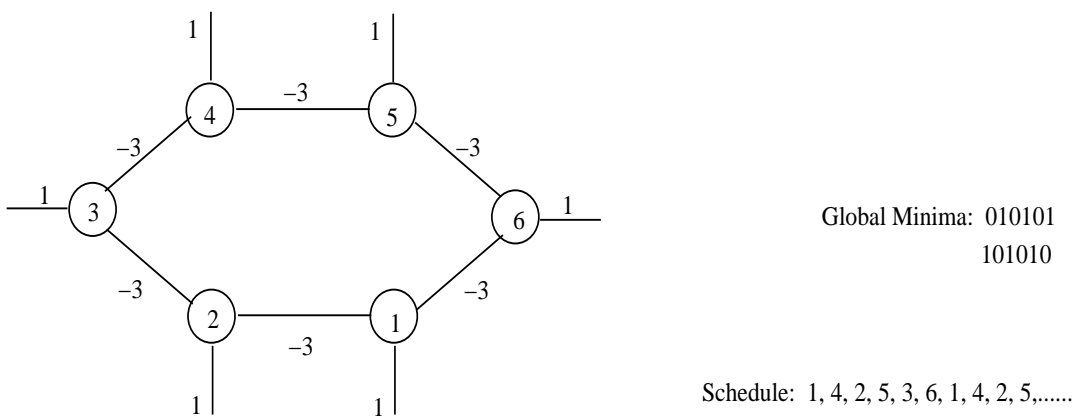

Figure 8: No uniform algorithm exists that guarantees to optimize rings even under a central scheduler.

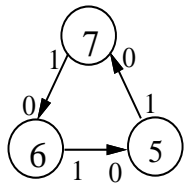

Figure 9: The uniform algorithm is not self-stabilizing in cyclic networks.

DEFINITION **4.1** A scheduler has the *fair exclusion* property if for every two neighbors, one is executed without the other infinitely often.

Intuitively, a distributed scheduler with fair exclusion will no longer generate infinite sequences of the pathological execution schedules used in the previous subsection to prove the negative results. Instead, it is guaranteed that from time to time, every two neighboring units will not execute together.

As an alternative, we might weaken the requirement on the uniformity of the algorithm (that all nodes execute the same procedure). An *almost uniform* algorithm is when all the nodes perform the same procedure except one node that is marked unique. In the almost uniform version of algorithm *activate*, the root of the tree is marked and executes the procedure of Section 3.4 as if all its neighbors are pointing to it; i.e., it constantly sets $P_i^j$ to zero.

THEOREM **4.3** *Algorithm* activate *of Section 3.4 has the following properties: 1. It converges to a global minimum and is self-stabilizing*[6] *in networks with tree-like topologies under a distributed scheduler with fair exclusion. 2. The algorithm also converges in tree-like subnetworks (but is not self-stabilizing) when the network has cycles. 3. It is self-stabilizing for any topology if an* almost *uniform algorithm is applied, even under a* pure *distributed scheduler.*

---

6. The initialization step of the algorithm is omitted in the self-stabilizing version.

For proof see appendix.

## 5. Extensions to Arbitrary Networks

The algorithm we presented in Section 3 is limited in that it is restricted to nodes of tree-like subnetworks only. Nodes that are part of a cycle execute the traditional activation function which may lead to the known drawbacks of local energy minima and slow convergence. In this section we discuss generalizations of our algorithm to nodes that are part of cycles, that will work well for near-tree networks. A full account of this extension is deferred for future work.

A well known scheme for extending tree algorithms to non-tree networks, is *cycle-cutset decomposition* (Dechter, 1990), used in Bayes networks and constraint networks. Cycle-cutset decomposition is based on the fact that an instantiated variable cuts the flow of information on any path on which it lies and therefore it changes the effective connectivity of the network. Consequently, when the group of instantiated variables cuts all cycles in the graph, (e.g., a cycle-cutset), the remaining network can be viewed as cycle-free and can be solved by a tree algorithm. The complexity of the cycle-cutset method can be bounded exponentially in the size of the cutset in each connected component of the graph (Dechter, 1992). We next show how to improve our energy minimization algorithm, *activate* using the cycle-cutset idea.

Recall that the energy minimization task is to find a zero/one assignment to the variables $X = \{X_1, ..., X_n\}$ that maximizes the goodness function. Define $Gmax(X_1, ..., X_n) = max_{X_1,...,X_n} G(X_1, ..., X_n)$. The task is to find an activation level $X_1, ..., X_n$ satisfying

$$Gmax(X_1, ...X_n) = max_{X_1,...,X_n}(\sum_{i<j} w_{i,j} X_i X_j + \sum_i \theta_i X_i). \qquad (2)$$

Let $Y = \{Y_1, ..., Y_k\}$ be a subset of the variables $X = \{X_1, ..., X_n\}$. The maximum can be computed in two steps. First compute the maximum goodness conditioned on a fixed assignment $Y = y$, then maximize the resulting function over all possible assignments to $Y$. Let $Gmax(X|Y = y)$, be the maximum goodness value of G conditioned on $Y = y$. Clearly,

$$Gmax(X) = max_{Y=y} Gmax(X|Y = y) = max_{Y=y} max_{\{X=x|x_Y=y\}}\{G(X)\},$$

where, $x_Y$ is the zero/one value assignments in the instantiation $x$ that are restricted to the variable subset $Y$. If the variables in $Y$ form a cycle-cutset, then the conditional maxima $Gmax(X|Y = y)$ can be computed efficiently using a tree algorithm. The overall maxima may be achieved subsequently by enumerating over all possible assignments to $Y$. Obviously, this scheme is effective only when the cycle-cutset is small. We next discuss some steps towards implementing this idea in a distributed environment.

Given a network with a set of nodes $X = \{X_1, ..., X_n\}$, and a subset of cutset variables $Y = \{Y_1, ..., Y_k\}$, presumably a cycle-cutset, and assuming a fixed, unchangeable assignment $Y = y$, the cutset variables behave like leaf nodes, namely, they select each of their neighbors as a parent if that neighbor does not point to them. Thus, a cutset variable may have several parents and zero or more child nodes.

Considering again the example network in Figure 3b and assuming node (7) is a cutset variable, a tree-directing may now change so that node (7) points both to (5) and to (6),

(6) points to (5) and (5) remains the root. Note that with this modification all arcs are directed and the resulting graph is an acyclic directed graph. Once the graph is directed, each regular non-cutset node has exactly the same view as before. It has one parent (or no parent) and perhaps a set of child nodes, some of which may be cutset nodes. It then computes goodness values and activation values almost as before.

An algorithm along these lines will compute the maximum energy conditioned on $Y = y$, if $Y$ is a cycle-cutset. Note however that such an assignment is not guaranteed to converge to a *local* maxima of the original (Hopfield) activation function. Some of the cutset nodes may be unstable relative to this function.

Enumerating all the conditional maxima to get a global maxima cannot be done distributedly, unless the cutset size is small. When the cutset is small, the computation can be done in parallel, yielding a practical distributed solution for networks, as follows. Once the tree-directing part is accomplished, a node computes a collection of goodness values, each indexed by a conditioning assignment $Y = y$. The goodness values of a node that are associated with the cutset assignment $Y = y$ will be computed using the goodness values of child nodes that are also associated with the same assignment $Y = y$. The maximum number of goodness values each node may need to carry is exponential in the cutset size. Upon convergence, the roots of the trees will select an assignment $Y = y$ that will maximize the overall goodness value and propagate this information down the tree so that nodes will switch values accordingly. The above algorithm is certainly not in the connectionist spirit and is practically limited to small cutsets. Its advantage is that it finds a true global optimum.

In the following subsection, we will modify the cutset approach more towards the connectionist spirit by integrating the cutset scheme with a standard energy minimizing activation function. This yields a connectionist-style algorithm with a simple activation function and limited memory requirements once the identity of the cycle-cutset nodes is known. We can determine the cutset variables initially using a centralized algorithm for computing a small cutset (Becker & Geiger, 1994). Although not guaranteed to find a global solution, the new activation function is more powerful than standard approaches on cyclic topologies.

## 5.1 Local search with cycle cutset

Algorithm *activate-with-cutset* in Figure 10 assumes that the cutset nodes are known a priori; this time however, their values are changing using standard local techniques (e.g., Hopfield). The algorithm is well-defined also when the cutset nodes do not cut all cycles or when the cutset is not minimal. However, it is likely to work best when the cutset is small and when it cuts all cycles.

Note that the goodness value computation of cutset nodes (step 4) is not performing the maximization operation over the two possible activation values of the cutset variables since the activation value of cutset nodes is fixed as far as the tree algorithm is concerned. Intuitively, if performed sequentially the algorithm would iterate between the following two steps: 1. finding a local maximum using Hopfield activation function for the cutset variables; 2. finding a global maximum conditioned on the cutset values determined in the previous step via the tree algorithm. In the connectionist framework these two steps are not synchronized. Nevertheless, the algorithm will converge to a local maxima relative to

---

**Algorithm activate-with-cutset (unit i)**
**Assumption: The cutset nodes are given a priori.**

1. Initialization: If first time, then $(\forall j)\ P_i^j = 0$;

2. Tree directing:
   If $i$ is a cutset node, for every neighbor $(j)$, if $P_j^i = 0$, then $P_i^j = 1$;
      (neighbors become parents unless they already point to it)
   else (not a cutset node), if there exists a single neighbor $k$, such that $P_k^i = 0$,
      then (part of a tree but not a root) $P_i^k = 1$ and for all other neighbors $j$, $P_i^j = 0$;
   else (root or non-tree node), for all neighbors $P_i^j = 0$;

3. Assigning activation values:
   If all neighbors of $i$ point to it except maybe one (i.e., it is part of a tree) then,
   $$X_i = \begin{cases} 1 & \text{if } \sum_j ((G_j^1 - G_j^0)P_j^i + w_{i,j}X_j P_i^j) \geq -\theta_i \\ 0 & \text{otherwise} \end{cases}$$
   else (a cutset node or a node that is not yet part of any tree), Compute Hopfield:
   $$X_i = \begin{cases} 1 & \text{if } \sum_j w_{i,j} X_j \geq -\theta_i \\ 0 & \text{otherwise} \end{cases}$$

4. Computing goodness values: (only nodes in trees need goodness values)
   If $i$ is a cutset node, then For each neighbor $j$,
   $G_i^0 = X_i \theta_i$,
   $G_i^{j1} = X_i(\theta_i + w_{ij})$    ($G_i^0, G_i^{j1}$ are goodness values for neighbor $j$).
   else (a regular tree node),
   $G_i^0 = max\{\sum_{j \in neighbors(i)} G_j^0 P_j^i, \sum_{j \in neighbors(i)} G_j^1 P_j^i + \theta_i\}$;
   $G_i^1 = max\{\sum_{j \in neighbors(i)} G_j^0 P_j^i, \sum_{j \in neighbors(i)} (G_j^1 P_j^i + w_{i,j} P_i^j) + \theta_i\}$;

---

Figure 10: Algorithm activate-with-cutset

Hopfield algorithm as well as a conditional global maxima relative to the cutset variables. Convergence follows from the fact that the tree directing algorithm is guaranteed to converge given fixed cutset variables. Once it does, a node flips its value either as a result of a Hopfield step or in order to optimize a tree. In both steps the energy does not increase.[7]

EXAMPLE **5.1** The following example demonstrates how the algorithm finds a better minimum than what is found by the standard Hopfield algorithm when there are cycles. Consider the energy function: $energy = 50AB - 200BC - 100AC - 3AD - 3DE - 3AE + 0.1A + 0.1B + 0.1C + 4E + 4D$. The associated network consists of two cycles: $A, B, C$ and $A, D, E$. If we select node $A$ as a cutset node, the network would then be cut into two acyclic (tree-like) subnetworks. Assume that the network starts with a setting of zeros ($A, B, C, D, E = 0$). This is a local minimum (energy = 0) of the Hopfield algorithm. Our activate-with-cutset algorithm breaks out of this local minimum by optimizing the acyclic subnetwork $A, B, C$ conditioned on $A = 0$. The result of the optimization is the assignment $A = 0, B = 1, C = 1, D = 0, E = 0$ with $energy = -199.7$. It is not a stable state because $A$ obtains an excitatory sum of inputs (50) and therefore flips its value to $A = 1$ using its Hopfield activation algorithm. The new state $A, B, C = 1, D, E = 0$ is also a local minimum of the Hopfield paradigm ($energy = -249.7$). However, since nodes $A, D, E$ form a tree, the activate-with-cutset algorithm also manages to break out of this local minimum. It finds a global solution conditioned on $A = 1$ which happens to be the global minimum $A, B, C, D, E = 1$ with $energy = -250.97$. The new algorithm was capable of finding the only global minimum of the energy function and managed to escape two of the local minima that trapped the Hopfield algorithm.

It is easy to see that algorithm *activate-with-cutset* improves on *activate* in the following sense:

THEOREM **5.1** *If $a_1$ is a local minimum generated by* activate *and $a_2$ is a local minimum generated by* activate-with-cutset, *then if $a_1$ and $a_2$ have the same activation value on all non-tree nodes then, $G(a_2) \leq G(a_1)$.*

## 5.2 Local search with changing cutset variables

We can imagine a further extension of the cutset scheme idea that will improve the resulting energy level further by conditioning and optimizing relative to many cutsets. In a sequential implementation the algorithm will move from one cutset to the next, until there is no improvement. This process is guaranteed to monotonically reduce the energy. It is unclear however how this tour among cutsets can be implemented in a connectionist environment. It is not clear even how to identify one cutset distributedly. Since finding a minimal cycle-cutset is NP-complete a distributed algorithm for the problem is unlikely to exist. Nevertheless, there could be many brute-force distributed algorithms that may find a good cutset in practice. Alternatively, cutset nodes may be selected by a process that randomly designates a node to be a cutset node.

---

7. Fluctuations that temporarily increase the energy may occur from time to time before the tree propagation has completely stabilized. For example, a node may use the goodness values of its children before their goodness values are ready (but after they point to the node).

In the following paragraphs we outline some ideas for a uniform connectionist algorithm that allows exploration of the cutset space. We propose the use of a random function to control the identity of the cutset nodes. The random process by which a node becomes a cutset node or switch from a cutset node to a regular node may be governed by a random heuristic function $f()$. A non-tree node may turn into a cutset node with probability $P = f()$. A cutset node may turn into a non-cutset node if it becomes part of a tree or by the random process with probability $P = g()$. The function $f()$ should be designed in a way that it will assign high probabilities to nodes with potential to become "good" cutset nodes. The probability of de-selecting a cutset node may be defined as $g() = 1 - f()$.

Algorithm *activate-with-cutset* can be augmented with a cutset selection function that will be running in parallel with the three procedures (tree-directing, assigning activation values and goodness computing). Thus, we may add a forth procedure that selects (or de-selects) the node as a cutset node with probability $P = f()$. Note that the randomly selected cutset is not perfect and that there might be too many or too few cutset nodes.

As long as there are cycles, cutset nodes should be selected. At the same time, nodes functioning too long as cutset nodes should be de-selected thus reducing the chances for redundant cutset nodes while continuously exploring the space of possible cutsets.

One way to implement a heuristic function $f$ is to base it on the following ideas: 1. Increase probability to non-tree nodes that have not been cutset nodes for a long time. 2. Increase probability to nodes that have not flipped their value for a long time. 3. Increase the probability to nodes with high connectivity.

Note that a de-selected cutset node may cause a chain reaction of undirecting nodes. Nodes that lost their tree-pointers become not-part-of-tree and thus have a potential to become cutset nodes. The network may continue the tour in the cutset space indefinitely and may never become static. The selection-de-selection process may never converge. Nevertheless, if a function $f$ is designed to allow enough time for convergence in between cutset changes than during the whole process, the energy tends to decrease. Temporary fluctuations may sometimes cause an energy increase when a node relies on its not yet stable neighbors.[8] We conjecture that such a heuristic function $f$ can be constructed so as to allow trees to be stabilized before they are distroyed by de-selection. Formalizing the algorithm's properties and further investigation and experimentation are left for future research.

## 6. Conclusions

The main contributions of the paper are:

1. We provide a connectionist activation function (algorithm activate, Figure 5), that is self-stabilizing and is guaranteed to converge to a global minima in linear time for tree-like networks. On general networks the algorithm will generate a global minima on all tree subnetworks and on the rest of the network it will coincide with regular local gradient activation functions (e.g., Hopfield). The algorithm dominates an arbitrary local search connectionist algorithm in the following sense: If $a_1$ is a local minimum generated by *activate* and $a_2$ is a local minimum generated by a corresponding local-search method, then if $a_1$ and $a_2$ have the same activation values on all non-tree nodes

---

8. For example, temporarily relying on old goodness values of a de-selected node.

(if it is a tree then the set is empty), then the energy of $a_1$ is smaller or equal to the energy of $a_2$.

2. We showed that *activate* can be further extended using the cycle-cutset idea. The extended algorithm called *activate-with-cutset* (Figure 10) is guaranteed to converge and generate solutions that are at least as good and normally better than algorithm *activate*. The algorithm converges to conditional global minima relative to the values of the cutset variables. If $a_1$ is a local minima generated by *activate* and $a_2$ is the local minima generated by *activate-with-cutset* then if $a_1$ and $a_2$ have the same activation values on all the cutset variables (if it is a tree, the cutset is empty) than the energy of $a_2$ is smaller or equal to the energy of $a_1$. Therefore *activate-with-cutset* is better than *activate* which in turn is better than a regular energy-minimization connectionist algorithm in the above sense. A third variation of the algorithm is sketched for future investigation. The idea is that the cutset nodes are randomly and continuously selected, thus allowing exploration of the cutset space.

3. We stated two negative results: 1) Under a pure distributed scheduler no *uniform* algorithm exists to globally optimize even simple chain-like networks. 2) No uniform algorithm exists to globally optimize simple cyclic networks (rings) even under a central scheduler. We conjecture that these negative results are not of significant practical importance since in realistic schedulers the probability of having infinite pathological scheduling scenarios approaches zero. We showed that our algorithm converges correctly (on tree-like subnetworks) when the demand for pure distributed schedulers is somewhat relaxed; i.e., adding either fair exclusion, almost uniformity or cycles. Similarly, self-stabilization is obtained in acyclic networks or when the requirement for a uniform algorithm is relaxed (almost uniformity).

The negative results apply to connectionist algorithms as well as to parallel versions of local repair search techniques. The positive results suggest improvements both to connectionist activation functions and to local repair techniques.

We conclude with a discussion of two domains that are likely to produce sparse, near-tree networks and thus benefit from the algorithms we presented: inheritance networks and diagnosis.

Inheritance is a straightforward example of an application where translations of symbolic rules into energy terms form networks that are mostly cycle free. Each arc of an inheritance network, such as $A$ ISA $B$ or $A$ HAS $B$ is modeled by the energy term $A - AB$. The connectionist network that represents the complete inheritance graph is obtained by summing the energy terms that correspond to all the ISA and HAS relationships in the graph. Nonmonotonicity can be expressed if we add penalties to arcs and use the semantics discussed by Pinkas (1991b, 1995). Nonmonotonic relationships may cause cycles both in the inheritance graph and the connectionist network (e.g. Penguin ISA Bird; Bird ISA FlyingAnimal; Penguin ISA not(FlyingAnimal)). Multiple inheritance may cause cycles as well, even when the rules are monotonic (e.g., Dolphin ISA Fish; Dolphin ISA Mammal; Fish ISA Animal; Mammal ISA Animal). Arbitrary constraints on the nodes of the graph may be introduced in this model. Constraints may be represented as proposition logic formulas and then translated into energy terms (Pinkas, 1991) potentially causing cycles. In a "pure" inheritance network that has no multiple inherited nodes and no nonmonotonic relationships, the network is cycle-free and can be processed efficiently by various algorithms. If we allow multiple inheritance, nonmonotonicity, or arbitrary propositional constraints, we may introduce cycles into the network that are generated. Nevertheless, it is reasonable to assume that in large practical inheritance domains cycles (multiple inheritance, nonmonotonicity and arbitrary constraints) are only scarcely introduced and the few that exist may be handled by our extension using the cycle-cutset idea.

Another potential application that will generate mostly cycle-free subnetworks is diagnosis. Here is a possible formulation of a diagnosis framework. Let $X_1, X_2, ...X_n$ be True(1)/false(0) propositions that represent symptoms and hypotheses. In a diagnosis application we may have diagnosis rules of the form: $(\alpha_1 X1, \alpha_2 X_2, ..., \alpha_m X_m \rightarrow \beta X)$. These rules announce that the symptoms $X_1, ..., X_m$ with importance factors $\alpha_1, ..., \alpha_m$, suggest the hypothesis $X$ with sensitivity $\beta$. A subset of the symptoms may be enough to suggest the hypothesis if the sum of the importance factors of the active symptoms is larger than the sensitivity $\beta$. Intuitively, the larger the sum of the factors, the larger the support for the hypothesis. The corresponding energy function for a diagnosis rule is $\sum_i^m -\alpha_i X_i X + \sum_i^m \alpha_i X_i + \beta X$. In addition, arbitrary propositional constraints may also be added, like $(X \rightarrow X_i)$ i.e., if the hypothesis $X$ holds, so does the symptom $X_i$. $(X_1 \rightarrow (\neg X_2 \wedge \neg X_3) \wedge X_2 \rightarrow (\neg X_1 \wedge \neg X_3) \wedge X_3 \rightarrow (\neg X_1 \wedge \neg X_2))$ i.e., only one of the propositions $X_1, X_2, X_3$ can be true (mutual exclusion). Any propositional logic formula is allowed and nonmonotonicity may be expressed using conflicting constraints (augmented with importance factors). Quadratic energy functions may be generated from arbitrary propositional constraints by introducing hidden variables (Pinkas, 1991).

Sparseness of such networks emerges as a result of assuming conditional independency of symptoms relative to their hypothesis. Independency assumptions of this kind (that makes computation tractable) are quite common in actual implementations of Bayes networks, influence diagrams (Pearl, 1988), and certainty propagation of rule-based expert systems (Shortliffe, 1976). When our knowledge base consists only of diagnosis rules (and maybe the corresponding $X \rightarrow X_i$ rules) and the symptoms are all independent of each other, there are no cycles in the network, and the tree algorithm converges to a global maximum in linear time. When we add dependent symptoms which affect a hypothesis through more than one path; e.g., $X_1 \rightarrow X$, and $X_1 \rightarrow X_2 \rightarrow ... \rightarrow X$, or when we start adding arbitrary constraints, cycles are added. When dependent symptoms and arbitrary constraints are only scarcely added, the network generated will most likely lend itself efficiently to the activate-with-cutset algorithm.

Abandonment of efficient algorithms exists for both inheritance and diagnosis in their tractable forms. Our algorithm offers both to solve efficiently tractable versions of the problem and to approximate intractable versions of it in massively parallel, simple to implement methods. The efficiency of the suggested process depends on the "closeness" of the problem to an ideal, tractable form.

## A. Appendix

*Proof sketch* of theorem 4.3: The second and third phases of the algorithm are adaptations of an existing dynamic programming algorithm (Bertelé & Brioschi, 1972), and their cor-rectness is therefore not proved here. The self-stabilization of these steps is obvious because no variables are initialized. The proof is therefore dependent on the convergence of the tree directing phase.

Let us first assume that the scheduler is distributed with fair exclusion and that the network is a tree. The first part of the theorem is proved by points 1-4. We want to show that the tree-directing algorithm converges, that it is self-stabilizing and that the final stable result is that the pointers $P_i^j$ represent a tree. Points 5 and 6 prove parts 2 and 3 of the theorem. A node is called *legal* if it is either a root (i.e., all its neighbors are legal, point to it and it doesn't point to any of them), or an intermediate node (i.e., it points to one of the neighbors and the rest of its neighbors are all legal and point back). A node is called a *candidate* if it is an illegal node and has all its neighbors but one pointing to it. We would like to show that:

1. The property of being legal is stable; i.e., once a node becomes legal it will stay legal.

2. A state where the number of illegal nodes is $k > 0$, leads to a state where the number of illegal nodes is less than $k$; i.e., the number of illegal nodes decreases and eventually all nodes turn legal.

3. If all the nodes are legal then the graph is marked as a tree.

4. The algorithm is self-stabilizing for trees.

5. The algorithm converges even if the graph has cycles (part 2 of the theorem).

6. The algorithm is self-stabilizing in arbitrary networks if an almost uniform version is used, even under a distributed scheduler (part 3 of the theorem).

We will now prove each of the above points.

1. Show that a legal state is stable. Assume a legal node $i$ becomes illegal. It is either a root node and one of its children became illegal, or an intermediate node whose one of its children became illegal (it cannot be that its parent suddenly points to $i$ or that one of the children stopped pointing and still is legal). Therefore, there must be a chain of $i_1, i_2, ..., i_k$ of nodes that became illegal. Since there are no cycles, there must be a leaf that was legal and turned illegal. This cannot occur since a leaf does not have children; leading to a contradiction.

2. Show that if there are illegal nodes, their number is reduced. To prove this claim we need three steps:

    (a) Show that eventually, if there are illegals, then there are also candidates. Because of the fair exclusion, eventually a state is reached where each node has been executed at least once. Assume that at least one node is illegal, and all the illegal nodes are not candidates. If a node is illegal and not a candidate, then either it is a root-type (all point to it) but at least one of its children is illegal, or there are at least two of its neighbors that are illegal. Suppose there are no root-type illegal nodes. Then all illegal nodes have at least two

illegal neighbors. Therefore there must be a cycle that connects illegal nodes (contradiction). Therefore, one of the illegal nodes must be root-type. Suppose $i$ is a root-type illegal node. It must have a neighbor $j$ which is illegal. Consider the subtree of $j$ that does not include $i$: it must contain illegal nodes. If there are no root-type illegal nodes we get a contradiction again. However, if there is a root-type node, we eliminate it and look at the subtree of some illegal $j'$ that does not include $j$. Eventually, since the network is finite, we obtain a subtree with no root-like illegal nodes but which includes other illegal nodes. This leads to a contradiction. The conclusion is that there must be candidates if there are illegal nodes.

(b) Show that a candidate is stable unless it becomes legal.

If a node $i$ is a candidate, all its legal children remain legal. There are three types of candidate nodes (node $j$ is an illegal neighbor of $i$):

  i. node $j$ points to $i$;
  ii. the pointer goes in both directions;
  iii. there is no pointer from $i$ to $j$ or vice-versa.

All possible changes in the pointers $P_i^j$ or $P_j^i$ will cause $i$ to remain a candidate or to turn legal (the rest of the pointers will not be changed).

(c) Show that every candidate node will eventually turn legal: Assume $j$ is the illegal neighbor of the candidate $i$. In the next execution of $i$ without $j$ (fair exclusion), if $P_j^i = 0$ then $i$ becomes legal by pointing to $j$; otherwise, $i$ becomes a root-type candidate (all its neighbors point to it) but $j$ is illegal. We will prove now that if an illegal node $j$ points to $i$ then eventually a state is reached where either $j$ is legal or $P_j^i = 0$, and that this proposition is stable once it holds. If this statement is true then when $i$ is executed eventually, if $j$ is legal then all of $i$'s neighbors are legal and therefore $i$ turns legal. If $j$ is illegal then $P_j^i = 0$, and $i$ will point to it ($P_i^j = 1$) making itself legal.

We next prove that if $j$ is an illegal node pointing to $i$ then there will be a state where either $j$ is legal or $P_j^i = 0$, and this state is stable. We prove it by induction on the size of the subtree of $j$ that does not include $i$.

Base step: If $j$ is a leaf and $j$ points to $i$ then if at the time $j$ is executed (without $i$) $P_i^j = 0$, then node $j$ points to $i$ and becomes legal; otherwise, $j$ updates $P_j^i = 0$. This status is stable because the legal state is stable and since a leaf will point to a node only if it turns legal.

Induction step: Assume hypothesis is True for trees of size less than $n$. Suppose $j$ is the illegal neighbor if $i$. Node $j$ points to $i$ and it has $j_1, ..., j_k$ other neighbors. Because we assume that all nodes were executed at least one time, since $j$ points to $i$ we assume that at the last execution of $j$ all the other neighbors $j_1, ..., j_k$ pointed to $j$. The subtrees rooted by $j_l$ (not including $j$) are of size less than $n$ and therefore by the hypothesis there will be a state where all the nodes $j_1, ..., j_k$ are either legal or $P_{j_l}^j = 0$. This state is stable, so when eventually $j$ is executed, it will either point to $i$ turning legal (if all $j_1, ..., j_k$ are pointing to it), or it will make $P_j^i = 0$ (if some of its neighbors do not point to it). Since the status of

244

$j_1, ..., j_k$ is stable at that point, whenever $j$ is executed it will either become legal or its pointers become zero.

3. Show that if all the nodes are legal then the graph is marked as a tree: If a node is legal, then all its children are legal and point to it. Therefore each node represents a subtree (if not a leaf) and has one parent at the most. To show that there is only one root we make the following argument. If several roots exist, then because of connectivity, there is one node that is shared between at least two subtrees and therefore has two parents (contradiction).

4. The algorithm is self-stabilizing for cycle-free networks since no initialization is needed (in the proof we haven't use the first initialization step; i.e., $P_i^j = 0$). In the case where no cycles exist we do not need this step. The pointers can get any initial values and the algorithm still converges.

5. The algorithm (with $P_i^j = 0$ initialization) converges even if the graph has cycles. Since all the nodes start with zero pointers, a (pseudo) root of a tree-like subnetwork will never point toward any of its neighbors (since it is part of a cycle and all of its neighbors but one must be legal).

6. Show that the algorithm is self-stabilizing in arbitrary networks if an almost uniform version is used, even under a distributed scheduler. We need to show that a candidate will eventually turn legal even if its neighbors are executed in the same time.
Suppose node $i$ is a candidate and node $j$ is its illegal neighbor:

   (a) If $j$ is a root, then it will never point to $i$, and therefore $i$ will eventually turn legal by pointing to $j$.

   (b) If $i$ is the root, then $P_i^j = 0$, and if $j$ becomes legal it will point to $i$ making $i$ legal. Node $j$ will turn eventually legal using the following induction (on the size of the subtree of $j$):
   Hypothesis: In a subtree without a node that acts as a root, all illegal nodes will eventually turn legal.
   Base step: If $j$ is a leaf, it will point eventually to its neighbor $i$ which in its turn will make $j$ legal by $P_i^j = 0$.
   Induction step: If $j_1, ..., j_k$ are other neighbors of $j$, then they will eventually turn legal (induction hypothesis) while pointing to $j$. Eventually $j$ is executed and also turns legal.

   (c) Suppose neither $i$ nor $j$ are roots, but one of them is not part of a cycle (and therefore is part of a subtree that does not include a node marked as a root). Using the above induction, all the nodes in the subtree will eventually turn legal. As a result either $i$ or $j$ eventually turns legal, and therefore $i$ will eventually turn legal as well.

□

## 7. Acknowledgement

This work was supported in part by NSF grant IRI-9157636, by Air Force Office of Scientific Research, AFOSR 900136, by Toshiba of America and by a Xerox grant. We would also like to thank Kalev Kask for commenting on the latest version of this manuscript, Kaoru Mulvihill for drawing the figures, for Lynn Haris for editing and the anonymous reviewers who helped improve the final version of this paper. A shorter version of this paper appears earlier (Pinkas & Dechter, 1992).

## References

Ballard, D. H., Gardner, P. C., & Srinivas, M. A. (1986). Graph problems and connectionist architectures. Tech. rep. 167, University of Rochester.

Becker, A., & Geiger, D. (1994). Approximation algorithms for loop cutset problems. In *Proceedings of the 10th conference on Uncertainty in Artificial Intelligence (UAI-94)*, pp. 60–68 Seattle, Washington.

Bertelé, U., & Brioschi, F. (1972). *Nonserial Dynamic Programming*. Academic Press, New York.

Brandt, R. D., Wang, Y., Laub, A. J., & Mitra, S. K. (1988). Alternative networks for solving the traveling salesman problem and the list-matching problem. *IEEE International Conference on Neural Networks*, *2*, 333–340.

Collin, Z., Dechter, R., & Katz, S. (1991). On the feasibility of distributed constraint satisfaction. In *Proceedings of IJCAI* Sydney.

Dechter, R. (1990). Enhancement schemes for constraint processing: Backjumping, learning and cutset decomposition. *Artificial Intelligence*, *41(3)*, 273–312.

Dechter, R. (1992). Constraint networks. In *Encyclopedia of Artificial Intelligence, 2nd ed.*, pp. 276–285. John Wiley & Sons, Inc.

Dechter, R., Dechter, A., & Pearl, J. (1990). Optimization in constraint networks. In in R.M. Oliver, J. S. (Ed.), *Influence diagrams, belief nets and decision analysis*. John Wiley and Sons.

Feldman, J. A., & Ballard, D. H. (1982). Connectionist models and their properties. *Cognitive Science 6*.

Hinton, G., & Sejnowski, T. (1986). Learning and re-learning in boltzmann machines. In *Parallel Distributed Processing: Explorations in The Microstructure of Cognition I, in J. L. McClelland and D. E. Rumelhart*, pp. 282–317. MIT Press, Cambridge, MA.

Hopfield, J. J. (1982). Neural networks and physical systems with emergent collective computational abilities. In *Proceedings of the National Academy of Sciences 79*, pp. 2554–2558.

Hopfield, J. J. (1984). Neurons with graded response have collective computational properties like those of two-state neurons. In *Proceedings of the National Academy of Sciences 81*, pp. 3088–3092.

Hopfield, J. J., & Tank, D. W. (1985). Neural computation of decisions in optimization problems. *Biological Cybernetics, 52*, 144–152.

Kasif, S., Banerjee, S., Delcher, A., & Sullivan, G. (1989). Some results on the computational complexity of symmetric connectionist networks. Tech. rep. JHU/CS-89/10, Department of Computer Science, The John Hopkins University.

Korach, K., Rotem, D., & Santoro, N. (1984). Distributed algorithms for finding centers and medians in networks. *ACM Transactions on Programming Languages and Systems, 6(3)*, 380–401.

McClelland, J. L., Rumelhart, D. E., & Hinton, G. (1986). The appeal of pdp. In *J. L. McClelland and D. E. Rumelhart, Parallel Distributed Processing: Explorations in The Microstructure of Cognition I*. MIT Press, Cambridge, MA.

Minton, S., Johnson, M. D., & Phillips, A. B. (1990). Solving large scale constraint satisfaction and scheduling problems using a heuristic repair method. In *Proceedings of the Eighth Conference on Artificial Intelligence*, pp. 17–24.

Papadimitriou, C., Shaffer, A., & Yannakakis, M. (1990). On the complexity of local search. *ACM Symposium on the Theory of Computation*, 438–445.

Pearl, J. (1988). *Probabilistic Reasoning in Intelligent Systems: Networks of Plausible Inference*. Morgan Kaufmann Publishers, San Mateo, California.

Peterson, C., & Hartman, E. (1989). Explorations of mean field theory learning algorithm. *Neural Networks 2, 6*.

Pinkas, G. (1991). Energy minimization and the satisfiability of propositional calculus. *Neural Computation 3, 2*.

Pinkas, G., & Dechter, R. (1992). A new improved activation function for energy minimization. In *Proceedings of the Tenth National Conference on Artificial Intelligence (AAAI)*, pp. 434–439 San Jose.

Rumelhart, D. E., Hinton, G. E., & McClelland, J. L. (1986). A general framework for parallel distributed processing. In *in J. L. McClelland and D. E. Rumelhart, Parallel Distributed Processing: Explorations in The Microstructure of Cognition I*. MIT Press, Cambridge, MA.

Selman, B., Levesque, H., & Mitchell, D. (1992). A new method for solving hard satisfiability problems. In *Proceedings of the Tenth National Conference on Artificial Intelligence*, pp. 440–446.

Shortliffe, E. H. (1976). *Computer-Based Medical Consultation, Mycin*. Elsevier, New York.

Smolensky, P. (1986). Information processing in dynamical systems: Foundations of harmony theory. In *J. L. McClelland and D. E. Rumelhart, Parallel Distributed Processing: Explorations in The Microstructure of Cognition I*. MIT Press, Cambridge, MA.